\documentclass[letterpaper]{article} % DO NOT CHANGE THIS

\usepackage{aaai25}  % DO NOT CHANGE THIS
\usepackage{times}  % DO NOT CHANGE THIS
\usepackage{helvet}  % DO NOT CHANGE THIS
\usepackage{courier}  % DO NOT CHANGE THIS
\usepackage[hyphens]{url}  % DO NOT CHANGE THIS
\usepackage{graphicx} % DO NOT CHANGE THIS
\usepackage{natbib}  % DO NOT CHANGE THIS AND DO NOT ADD ANY OPTIONS TO IT
\usepackage{caption} % DO NOT CHANGE THIS AND DO NOT ADD ANY OPTIONS TO IT
\usepackage{algorithm}
\usepackage{xspace}
\usepackage{comment}
\usepackage{tabularx}
\usepackage{xcolor}        
\usepackage{enumitem}
\usepackage{algorithm}
\usepackage{algpseudocode}
\usepackage{lscape}
\usepackage{adjustbox}
\usepackage{tabularx}
\usepackage{graphicx}
\usepackage{float} 
\usepackage{subfigure}
\usepackage{color}
\usepackage[]{caption}
\usepackage{multirow}

\usepackage[utf8]{inputenc}
\usepackage[T1]{fontenc}
\usepackage{url}
\usepackage{booktabs}
\usepackage{amsfonts}
\usepackage{nicefrac}
\usepackage{microtype}
\usepackage{natbib}
\newcommand{\sysname}{AdaSkip\xspace}

\usepackage{newfloat}
\usepackage{listings}
\DeclareCaptionStyle{ruled}{labelfont=normalfont,labelsep=colon,strut=off} % DO NOT CHANGE THIS
\floatstyle{ruled}
\newfloat{listing}{tb}{lst}{}
\floatname{listing}{Listing}
\nocopyright %-- Your paper will not be published if you use this command
\title{\sysname: Adaptive Sublayer Skipping for Accelerating Long-Context LLM Inference}
\author{
    Zhuomin He\textsuperscript{\rm 1}\equalcontrib\thanks{Work done during their internship at Huawei Cloud.},
    Yizhen Yao\textsuperscript{\rm 1}\equalcontrib\footnotemark[\value{footnote}],
    Pengfei Zuo\textsuperscript{\rm 2}\equalcontrib,
    Bin Gao\textsuperscript{\rm 3}\footnotemark[\value{footnote}],
    Qinya Li\textsuperscript{\rm 1}\thanks{Corresponding author.},
    Zhenzhe Zheng\textsuperscript{\rm 1},
    Fan Wu\textsuperscript{\rm 1}
}
\affiliations{
    \textsuperscript{\rm 1}Shanghai Jiao Tong University\\
    \textsuperscript{\rm 2}Huawei Cloud\\
    \textsuperscript{\rm 3}National University of Singapore\\
    \{dean\_hzm, 1975275148\}@sjtu.edu.cn, pengfei.zuo@huawei.com, bingao@comp.nus.edu.sg\\
    \{qinyali, zhengzhenzhe\}@sjtu.edu.cn, fwu@cs.sjtu.edu.cn 
}

\usepackage{bibentry}
\begin{document}

\maketitle

\thispagestyle{plain} % 明确指定第一页使用plain样式

\begin{abstract}
Long-context large language models (LLMs) inference is increasingly critical, motivating a number of studies devoted to alleviating the substantial storage and computational costs in such scenarios. Layer-wise skipping methods are promising optimizations but rarely explored in long-context inference. We observe that existing layer-wise skipping strategies have several limitations when applied in long-context inference, including the inability to adapt to model and context variability, disregard for sublayer significance, and inapplicability for the prefilling phase. This paper proposes \sysname, an adaptive sublayer skipping method specifically designed for long-context inference. \sysname adaptively identifies less important layers by leveraging on-the-fly similarity information, enables sublayer-wise skipping, and accelerates both the prefilling and decoding phases. The effectiveness of \sysname is demonstrated through extensive experiments on various long-context benchmarks and models, showcasing its superior inference performance over existing baselines.
\end{abstract}


\section{Introduction}
Recently, large language models (LLMs) evolve to support long-context inference~\cite{xiao2024infllmtrainingfreelongcontextextrapolation, srivatsa2024prebleefficientdistributedprompt, deepseekai2024deepseekv2strongeconomicalefficient} up to 1M~\cite{liu2024worldmodelmillionlengthvideo,ai2024yiopenfoundationmodels}, unlocking more complex real-world applications such as personal agent~\cite{park2023generativeagentsinteractivesimulacra, Wang_2024}, document summarization ~\cite{wu2023vcsumversatilechinesemeeting}, and coding assistance~\cite{liu2023repobenchbenchmarkingrepositorylevelcode,bairi2023codeplanrepositorylevelcodingusing,jimenez2024swebenchlanguagemodelsresolve}. Long-context inference introduces more computational and storage demands. It is crucial to reduce the inference cost for long sequences.

Layer-wise skipping strategies, as an emerging technology, show great promise to reduce the LLM inference cost and latency by omitting the execution of transformer layers at specific positions, e.g., early skipping~\cite{del2023skipdecode, zhu2024hierarchicalskipdecodingefficient}, periodic skipping~\cite{liu2024accelerating}, and early exit~\cite{varshney2023accelerating, fan2024not, chen2024eellmlargescaletraininginference}.

However, we observe that these layer-wise skipping strategies all have their limitations in taking effect in long-context inference due to the following reasons.
First, existing layer-wise skipping strategies lead to a significant degradation in the generation quality due to predetermined fixed layers being skipped regardless of model and context variance. 
We observe that the importance distributions of transformer layers are different across models and contexts, and none of these strategies can perform consistently best across all models and contexts. 
Second, existing skipping strategies perform skipping at monolithic transformer layers which leads to suboptimal performance. We observe that the importance distributions of sublayers, i.e., attention and FFN modules, are independent. Moreover, in long-context inference, attention sublayers contribute significantly to inference latency~\cite{tang2024questqueryawaresparsityefficient, jiang2024minference10acceleratingprefilling}, highlighting the importance of prioritizing the skipping of more attention sublayers.
Third, existing layer-wise skipping strategies are limited to the decoding phase, neglecting optimization of the prefilling phase in long-context inference, where the latency of the prefilling phase, i.e., time to first token (TTFT), imposes a significant burden on long-context inference latency.

To address the above limitations, we propose \textit{\sysname}, an auto-adaptive, sublayer-wise skipping strategy tailored for long-context inference, which can benefit both the prefilling and decoding phases. 
Firstly, \sysname exploits on-the-fly similarity information during execution to adaptively identify the least important layers in different models, thereby improving the generation quality. Secondly, \sysname independently determines the importance distribution residing within sublayer modules like attention and FFN, enabling the sublayer-wise skipping. Finally, \sysname identifies the 
least important sublayers during both prefilling and decoding phases, significantly reducing the time and memory overhead of long-context scenarios. The code is released on Github\footnote{\url{https://github.com/ASISys/AdaSkip}}.

In summary, our contributions are as follows:
\begin{enumerate}
\item We perform a comprehensive analysis of the importance distributions of various components including layer and sublayer modules across a range of different models. Based on the analysis, we present the limitations of the existing layer-wise skipping strategies in accelerating long-context inference. 
\item We propose an auto-adaptive, sublayer-wise skipping strategy that works for both the prefilling and decoding phases in long-context scenarios.
\item We conduct extensive experiments on various long-context benchmarks and models, demonstrating that \sysname exhibits favorable inference performance over the baselines.
\end{enumerate}

\section{Background and Motivation}
In this section, 
we first perform a comprehensive exploration of the importance metric of the layer and sublayer-wise modules, then present observations on the characteristics of the importance distribution and motivate our design principles.

\subsection{IO Similarity and Transformer Module Importance} \label{sec:motivation:similarity}
We first define the metric, \textit{similarity}, to evaluate the importance of transformer layers and sublayer modules.
Given two $n$-dimensional vectors, $\vec{a}$ and $\vec{b}$, we characterize the cosine similarity between these vectors as their similarity, defined as follows:  %%如果a b是个vector，用vector的标识 $\vec{a}$
\begin{equation}
    \label{equ:cos_sim}
    \textit{Similarity($\vec{a}$, $\vec{b}$)} = \frac{\vec{a} \cdot \vec{b}}{\Vert \vec{a} \Vert \Vert \vec{b} \Vert}  = \frac{\sum_{i=1}^n a_ib_i}{ \sqrt{\sum_{i=1}^{n} a_i^2 }\sqrt{\sum_{i=1}^{n} b_i^2}}
\end{equation}

Following the existing works~\cite{liu2023deja,jaiswal2024ffn,fan2024not}, the similarity between the input and output (IO) vectors of the transformer module, i.e., IO similarity, can be used to evaluate the importance of a transformer module. Specifically, following the forwarding of each module, if the input vector of the module closely resembles the output vector, it indicates that the module contributes minimally to the forward propagation process. In other words, the current module contributes less \textit{importance} in terms of execution. Conversely, the current module possesses higher \textit{importance} in terms of execution if the IO similarity is low.

We further empirically validate the correlation between the IO similarity and the importance of a transformer module. Given an inference task, we conduct a first-round inference process to profile the IO similarity of each transformer layer. Subsequently, we execute a second-round inference process that selectively skips the layers based on varying degrees of the profiled IO similarity. Then we assess the quality of generated output by evaluating its GPT score~\cite{varshney2023accelerating, jaiswal2024ffn}.
%The GPT score leverages ChatGPT4 as a critic, utilizing a specialized prompt to direct GPT in assessing text quality on a scale from 1 to 10, where higher scores indicate superior generation quality.
The LeastSkip strategy, which skips the layers exhibiting the lowest IO similarity, experiences a substantial degradation in the GPT score (dropping below 1.0 even with one skipped layer), compared to the MostSkip strategy, which skips the layers with the highest IO similarity and yields GPT scores of 8.9, 6.1, and 4.2 when skipping 1, 3, and 5 layers, respectively.

\subsection{Existing Layer-wise Skipping Strategies} \label{sec:motivation:layerskip}
\begin{figure*}[t]
    \centering 
    \includegraphics[width=0.9\textwidth]{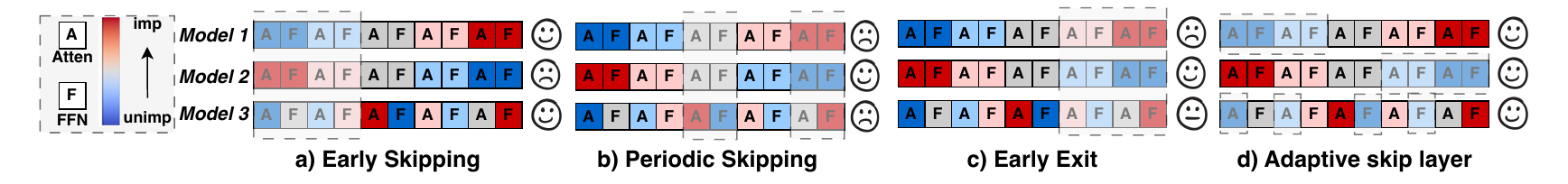}
    \caption{The comparisons of different skipping strategies. The dashed box indicates the layer to be skipped.}
    \label{fig:cmp}
\end{figure*}

Existing layer-wise skipping strategies propose skipping fixed layers with certain preferences to reduce inference execution time.
As shown in Figure~\ref{fig:cmp}, according to the strategies to skip layers, existing layer-wise skipping strategies can be broadly categorized into three types: early skipping~\cite{del2023skipdecode}, periodic skipping~\cite{liu2024accelerating}, and early exit~\cite{schuster2022confident, varshney2023accelerating, fan2024not, bae2023fast}. Early skipping~\cite{del2023skipdecode} always skips the first few layers that are predetermined. Early skipping can support batching operations but may skip the important layers. Periodic skipping~\cite{liu2024accelerating} periodically skips a few middle layers. It follows a predetermined frequency to skip one layer every several layers. Periodic skipping supports batching operations but cannot capture the varying importance of different layers. Early exit~\cite{varshney2023accelerating, fan2024not} always skips the last few layers. It evaluates whether the conditions (e.g., confidence level) are met after finishing the computation of each layer and the execution immediately exits upon condition fulfillment. Early exit may overlook the important layers that come later. Moreover, existing early exit strategies need to pay additional efforts and costs to either train classifier~\cite{del2023skipdecode} or fine-tune the model to counterbalance the information loss resulting from imperfect layer skipping~\cite{liu2024accelerating, varshney2023accelerating, fan2024not}.

\subsection{Motivation} \label{sec:motivation}
This subsection analyzes the limitations of existing LLM acceleration strategies for long-context inference.

\begin{figure*}[t]
    \centering 
    \subfigure[Vicuna-7B-16k]{
        \label{OPT}
        \includegraphics[width=0.3\textwidth]{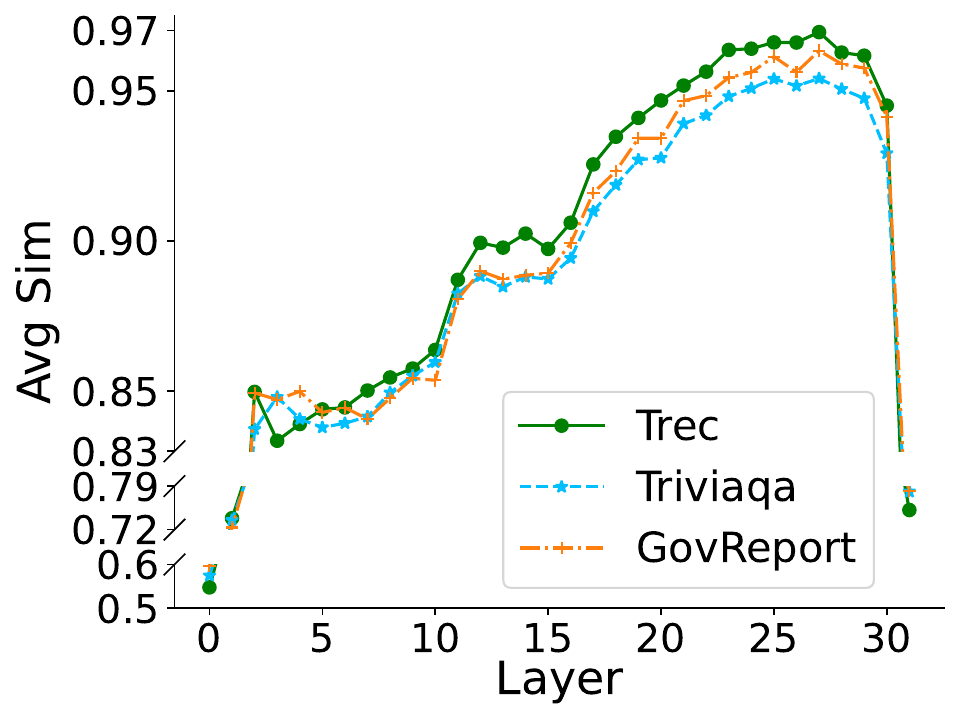}}
    \subfigure[InternLM-7B-8k]{
        \label{Falcon}
        \includegraphics[width=0.3\textwidth]{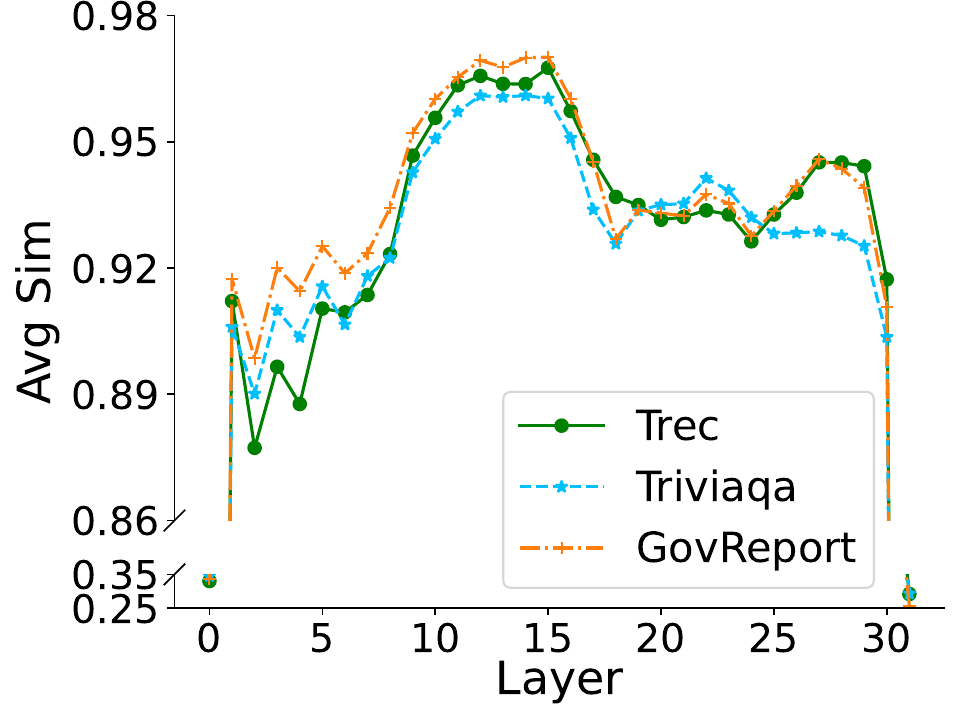}}
    \subfigure[LLaMA3.1-8B-128k]{
        \label{llama}
        \includegraphics[width=0.3\textwidth]{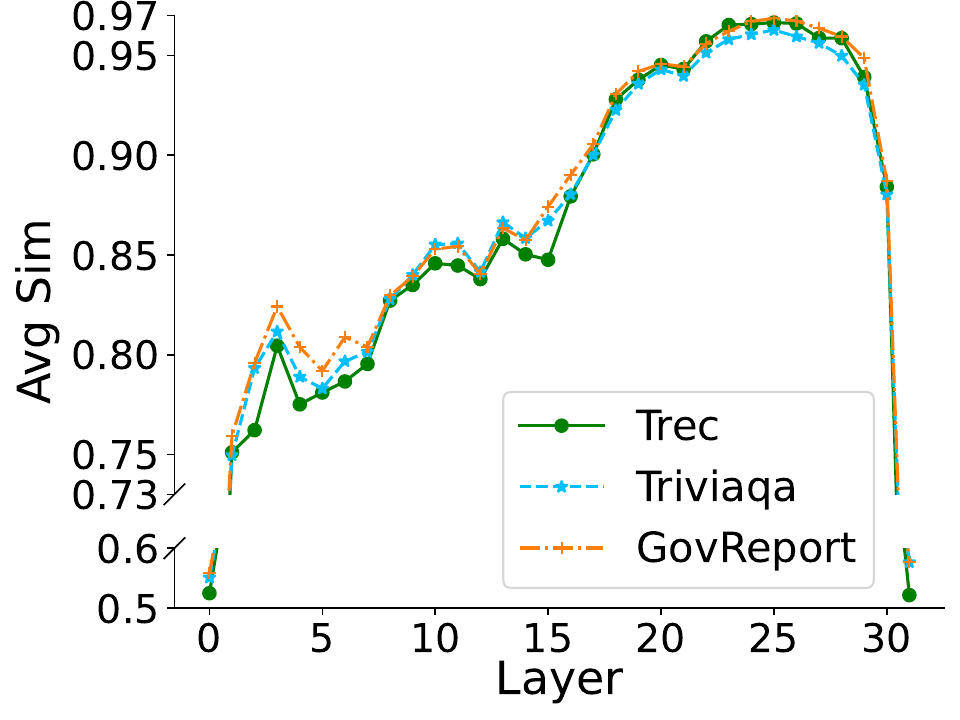}}
        \caption{IO similarities of different layers in various transformer models.}
        \label{cos_trans}
\end{figure*}

\textbf{Observation 1: \textit{The layer importance distribution exhibits significant variation across diverse models.}}
\label{sec:motivation:diffmodel}
We follow the same way used in the previous section to investigate the IO similarities of different layers on various models, in both prefilling and decoding phases. Figure~\ref{cos_trans} shows significant variation in the IO similarities of transformer layers for different models in three long-context datasets. Taking InternLM-7B-8k and LLaMA3.1-8B-128k as examples, layers with high IO similarity in InternLM-7B-8k appear in the middle, such as layers 12,13,14, and the curve is more irregular. Whereas layers with high IO similarity in LLaMA3.1-8B-128k, appear towards the end, with layers 27, 25, 28, 29, and 26 being the top 5 layers, and the curve is approximately monotonically ascending. This suggests that layer importance distributions vary among different models.
Existing layer-wise skipping strategies tend to consistently skip fixed layers, overlooking the differences in importance distribution across models, which restricts their adaptability to various models.  
Adaptive skipping strategies matching various models are required.  

\begin{figure*}[t]
    \centering 
    \subfigure[Vicuna-7B-16k]{
        \label{OPT}
        \includegraphics[width=0.3\textwidth]{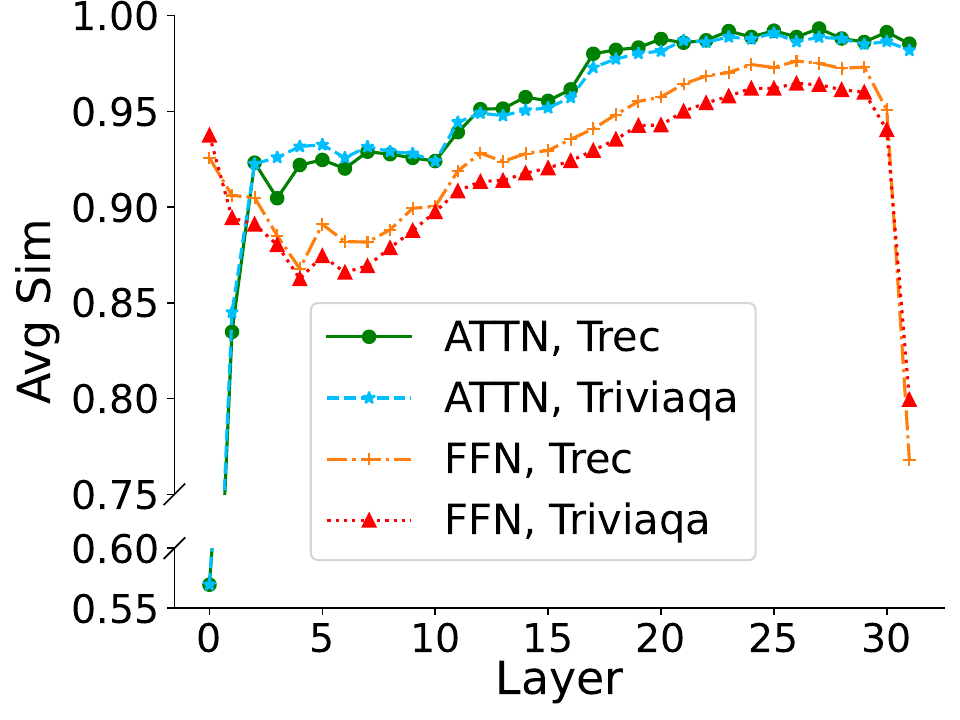}}
    \subfigure[InternLM-7B-8k]{
        \label{Falcon}
        \includegraphics[width=0.3\textwidth]{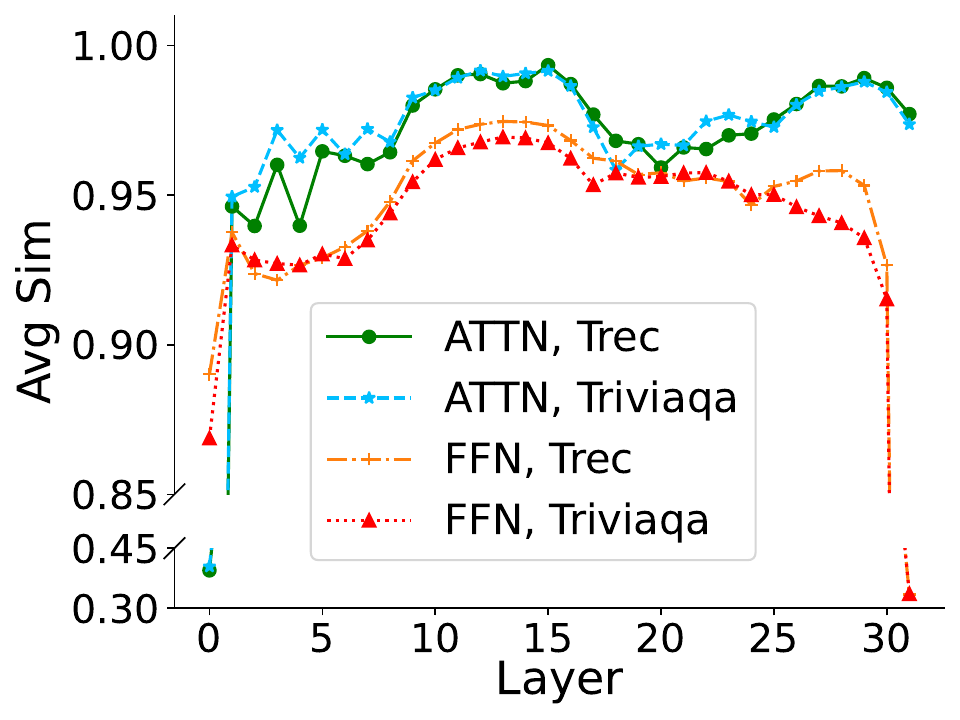}}
    \subfigure[LLaMA3.1-8B-128k]{
        \label{llama}
        \includegraphics[width=0.3\textwidth]{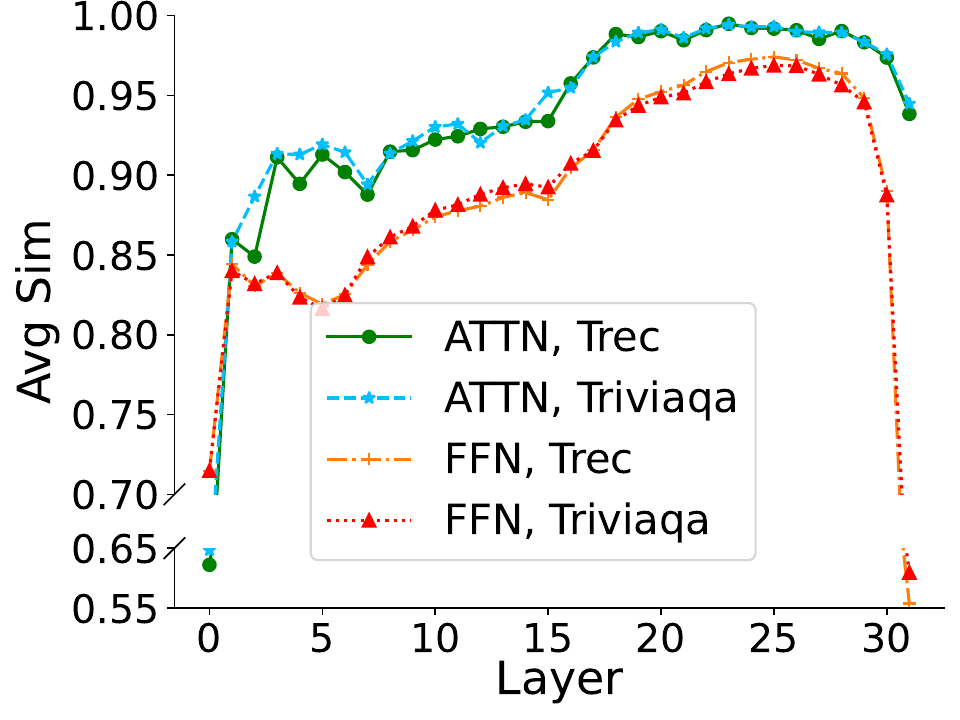}}
        \caption{IO similarities of attention (ATTN) and FFN modules in different layers.}
        \label{cos_atten_FFN}
\end{figure*}

\textbf{Observation 2: \textit{The importance distributions of attention and FFN modules are different.}}
\label{sec:motivation:sublayer}
We study the IO similarities of the sublayer-wise modules, i.e., attention and FFN.
As shown in Figure~\ref{cos_atten_FFN}, the sublayer-wise modules show diverse IO similarity distributions. 
Taking LLaMA3.1-8B-128k as an example, in the last 11 layers, the average IO similarity of attention is consistently around 0.97, indicating a high IO similarity. However, the highest average IO similarity of FFN in the last 11 layers is only 0.95, and it is relatively scattered. Furthermore, compared to FFN, attention modules demonstrate higher and more concentrated similarity, implying that a greater number of attention modules can be skipped, with the potential to save more KV cache in long-context inference.
The different characteristics in IO similarity distributions of attention and FFN suggest that the existing layer-wise skipping methodologies that monolithically skip entire transformer layers are sub-optimal. Consequently, the attention sublayer and FFN sublayer within one transformer layer should be considered separately.

\begin{figure*}[t]
    \centering 
    \subfigure[Vicuna-7B-16k]{
        \label{OPT}
        \includegraphics[width=0.3\textwidth]{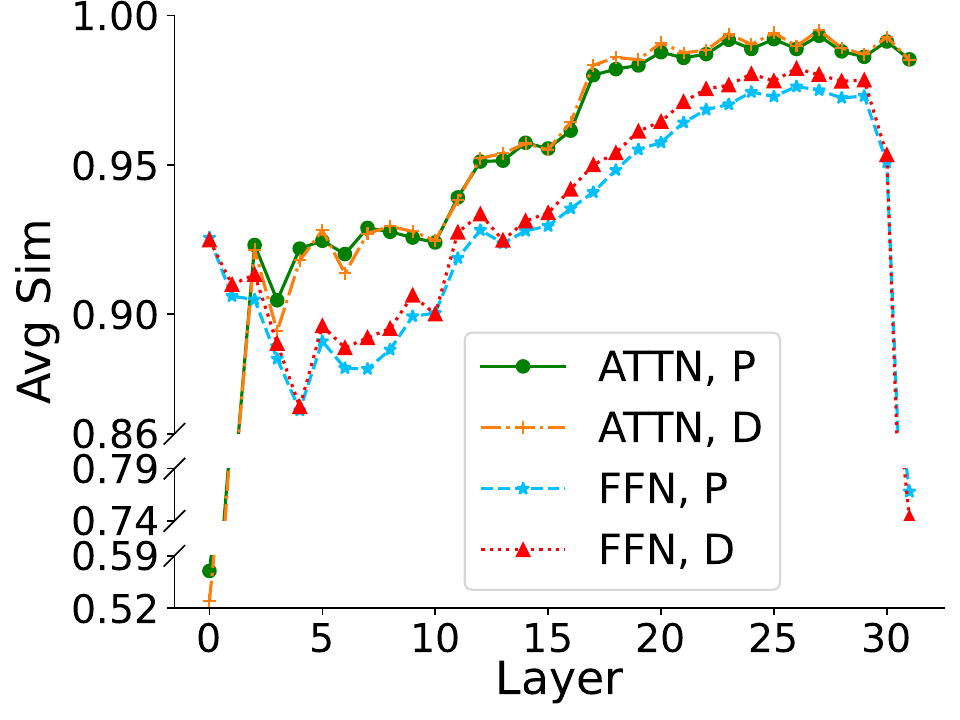}}
    \subfigure[InternLM-7B-8k]{
        \label{Falcon}
        \includegraphics[width=0.3\textwidth]{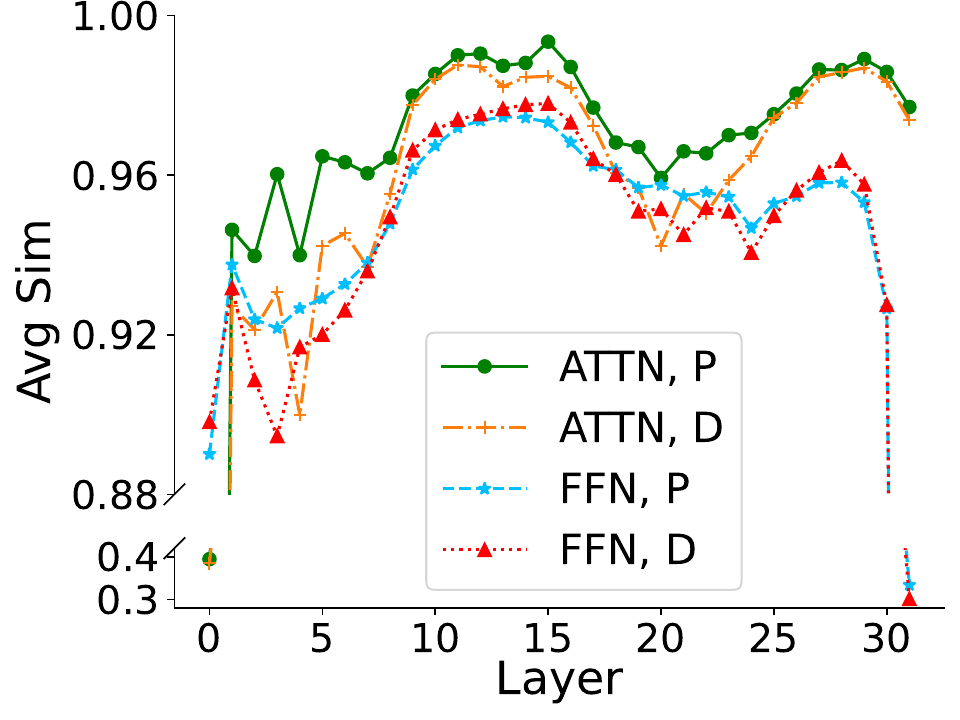}}
    \subfigure[LLaMA3.1-8B-128k]{
        \label{llama}
        \includegraphics[width=0.3\textwidth]{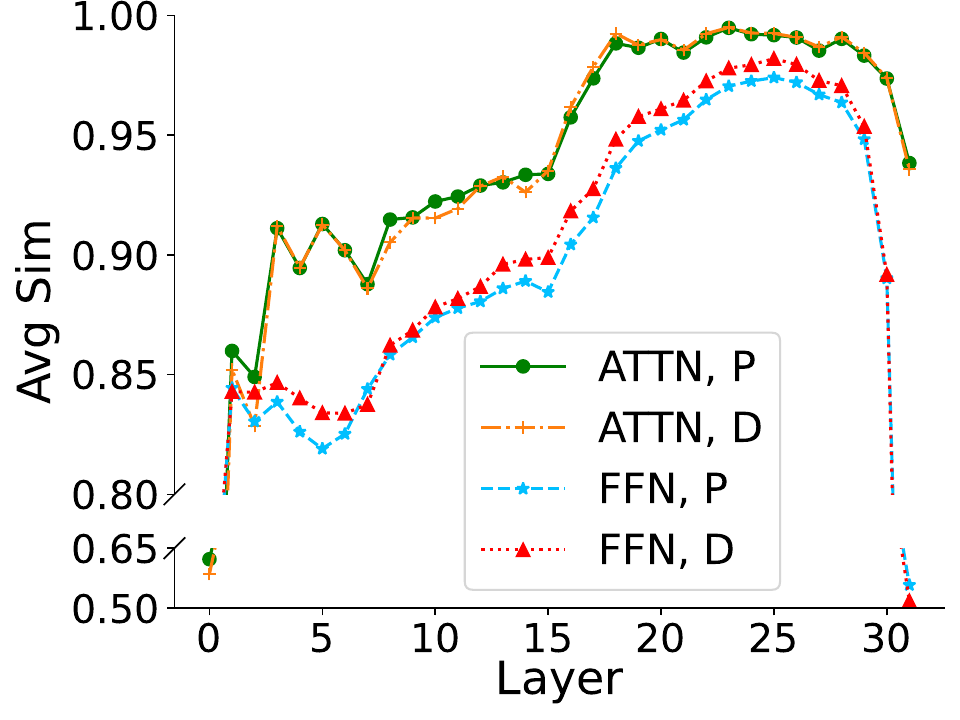}}
        \caption{IO similarities of sublayer modules in prefilling (P) and decoding (D) phases.}
\label{cos_pre_dec}
\end{figure*}

\textbf{Observation 3: \textit{The importance distribution of sublayers in the prefilling and decoding phases have similar trends but different fluctuation degrees.}} \label{sec:motivation:prefilling}
We further investigate the IO similarities of sublayer modules in the prefilling and decoding phases respectively. As shown in Figure~\ref{cos_pre_dec}, both attention and FFN sublayers display a consistent IO similarity trend between the prefilling and decoding phases, indicating that similar skipping strategies can be shared between the two phases.
What's more, we found a phenomenon that among all three models, each FFN sublayer has a higher IO similarity in the decoding phase than in the prefilling phase, which is different from that of attention sublayers. This suggests that we have the opportunity to skip more FFN sublayers in the decoding phase without affecting the model performance.

\textbf{Challenges.} \label{sec:challenges}
Based on the above observations, an efficient skipping strategy for long-context inference should have the following capabilities: (1) adaptability to various models, (2) independent decision-making for sublayer-wise skipping, and (3) the ability to skip the most unimportant layers in both the prefilling and decoding phases.

However, implementing such a skipping strategy encounters several challenges. First, limited prior information is available to guide the skipping decisions throughout the prefilling phase. Second, distinguishing the unique information corresponding to specific models and contexts, required for making adaptive choices, is far from straightforward.

\section{Methodology}\label{sec:method}
\subsection{Overview}

To tackle the above challenges, we propose a novel skipping strategy for long-context inference, called \sysname, which adaptively selects sublayer-wise modules to skip considering the characteristics of models and inference context. Specifically, \sysname efficiently learns the importance distributions from the past inference execution to construct the skipping strategy for the prefilling phase. It further improves the skipping decision by online importance learning from on-the-fly intermediate data during the decoding phase. By integrating the above techniques, \sysname can accurately skip the least important sublayer-wise modules, avoiding the mismatch of layer importance and layer skipping decisions in fixed layer-wise skipping strategies.

\subsection{Sublayer Skipping during Prefilling with Offline Importance Learning}
\label{sec:method:offline}

It is necessary to skip layers in prefilling phases during long-context inference, since the prefilling phase results in unacceptably high TTFT and substantial KV cache demands. However, existing layer-skipping strategies rarely consider skipping strategies in such phases. Moreover, since different models exhibit various similarity distributions, current fixed layer-skipping strategies cannot achieve optimal results.
The primary obstacle in devising an adaptive sublayer-wise skipping approach for the prefilling phase lies in the absence of prior knowledge before execution. To address this challenge, we propose an offline importance learning method that leverages the high correlation between historical prefilling features and new prefilling features. 

\textbf{Insight.}
\textit{Using sublayer-wise IO similarity feature from historical tasks can precisely predict the sublayer-wise skipping behavior for prefilling new inference tasks.} 
\begin{table}[t]
\centering
\setlength{\tabcolsep}{1mm}
\small  % 9pt
\begin{tabular}{cccc}
\toprule
\textbf{Type} & \textbf{Src} & \textbf{Dest} & \textbf{Layer Hit Rate} \\ 
\midrule
ATTN & TriviaQA & MFieldQA & 3.76/4, 4.86/6, 9.31/10\\
ATTN & MFieldQA & Wiki & 3.80/4, 5.54/6, 9.90/10\\
ATTN & TriviaQA & Wiki & 3.79/4, 5.50/6, 9.68/10\\
FFN & TriviaQA & MFieldQA & 3.66/4, 5.69/6, 9.56/10\\
FFN & MFieldQA & Wiki & 3.77/4, 5.97/6, 9.38/10\\
FFN & TriviaQA & Wiki & 3.75/4, 5.96/6, 9.64/10\\ 
\bottomrule
\end{tabular}
\caption{Average hit rate of unimportant layers using historical features across datasets in prefilling phases.}
\label{tab:transfer}
\end{table}
We perform the IO similarity analysis study of running inference tasks of multiple datasets~\cite{taori2023stanford} including 2WikiMQA, MultiFieldQA-en, and TriviaQA using LLaMA3.1-8B-128k and quantify the average hit rate of unimportant layers in the prefilling phase. We record the average IO similarity on the Src dataset in prefilling phases and test the hit rate on the Dest dataset.
% To be specific, "hit" indicates that the layers with the highest IO similarity, identified from historical prefilling tasks, match those layers with the highest IO similarity derived from the new prefilling task.
The results shown in Table~\ref{tab:transfer} reveal that historical IO similarity in prefilling phases gains a high hit rate for subsequent tasks, suggesting that this feature can be used in prediction and shared across different datasets.

\textbf{Method.}
\label{method:part1}
Based on the insight, the major workflow of offline importance learning consists of the similarity study and the corresponding deviation correction procedure.
Specifically, suppose $N$ inference tasks (samples) are used in offline importance learning. As for the inference task $T_i$ with prompt length $|T_i|$. Suppose that the model has $M$ transformer layers with $M$ attention sublayers and $M$ FFN sublayers. We first take notes of average similarity $\bar{Similarity}$ in the prefilling phase. The average similarity of the $j_{th}$ sublayer, $\bar{\textit{Similarity}_j}$, can be accumulated as: 
\begin{equation}
    \bar{\textit{Similarity}_j} = \frac{\sum^{N}_{i=1}\sum^{|T_i|}_{t=1} \textit{Similarity}(\vec{a}^j_{it}, \vec{b}^j_{it})}{\sum^{N}_{i=1} |T_i|}
\end{equation} 
where $\vec{a}^j_{it}$ and $\vec{b}^j_{it}$ are the input and output vectors of the $t$-th token in task $i$.
In addition, if the angle between vector $\vec{a}_{it}^j$ and $\vec{b}_{it}^j$ is not very large, the proportion of modulus of $\vec{a}_{it}^j$ and $\vec{b}_{it}^j$ relatively become prominent, suggesting some compensation needs to be applied.

However, due to the residual connections employed between each sublayer, the modulus of the input and output of one layer has minor variations, 
which implies that the average proportion of modulus can effectively compensate for the deviations. Hence, we use the average proportion of historical modulus of $\vec{a}_{it}^j$ and $\vec{b}_{it}^j$ in $j_{th}$ layer to scale $\vec{a}_{it}^j$ so that output vector $\hat{\vec{b}^j_{it}}$ is close to original $\vec{b}_{it}^j$. The average scale factor of $j$-th sublayer, $\bar{Scale_j}$, can be formulated as:
\begin{equation}
    \bar{\textit{Scale}_j} = \frac{\sum^{N}_{i=1}\sum^{|T_i|}_{t=1}\frac{\Vert \vec{b}^j_{it}\Vert}{\Vert \vec{a}^j_{it}\Vert}}{\sum^{N}_{i=1} |T_i|}
\end{equation}
we use $\bar{\textit{Scale}}_j$ to compensate the input $\vec{a}^j_{it}$, getting approximate output:
\begin{equation}
    \hat{\vec{b}^j_{it}}= \bar{\textit{Scale}_j} * {\vec{a}^j_{it}}
\end{equation}

After obtaining $\bar{\textit{Similarity}}$ and $\bar{\textit{Scale}}$ of each sublayer module, we sort all sublayers in descending order based on their $\bar{\textit{Similarity}}$, getting the sorted list $\textit{sorted}$ with $2M$ elements. Since there is a trade-off between the number of skipped layers and the generation quality, we introduce an acceleration ratio, $\alpha$, as a knob to control this trade-off. Given the acceleration ratio $\alpha$, the number of sublayers to be skipped, $m$, can be calculated as $m = M-\frac{M}{\alpha}$, and the targeted skipping sublayer number is $2m$. The top $2m$ sublayers in the $\textit{sorted}$ list are selected, forming the \textit{skipped} set.

\subsection{Extra FFN Sublayer Skipping during Decoding with Online Importance Learning} \label{sec:method:online}

\begin{table}[t]
\centering
\setlength{\tabcolsep}{1mm}
\small  % 9pt
\begin{tabular}{ccc}
\toprule
\textbf{Dataset} & \textbf{Size} & \textbf{Layer Hit Rate} \\ 
\midrule
TREC & 5 & 0.84/2, 2.67/4, 4.70/6\\
TREC & 20 & 1.08/2, 3.04/4, 4.90/6\\
TREC & 40 & 1.07/2, 3.09/4, 4.90/6\\
GovReport & 5 & 1.01/2, 2.94/4, 4.97/6\\
GovReport & 20 & 1.14/2, 3.01/4, 5.02/6\\
GovReport & 40 & 1.19/2, 3.03/4, 5.03/6\\
\bottomrule
\end{tabular}
\caption{Average hit rate of unimportant layers identified through different window sizes in the decoding phase.}
\label{tab:window-size}
\end{table}

Based on Observation 3, we find that regardless of attention or FFN sublayer, IO similarity of unimportant sublayers is always similar in prefilling and decoding phases, which suggests that we can reuse the layers selected from prefilling phases when decoding. What's more, we observe that each FFN sublayer has higher IO similarity in decoding phases compared with the prefilling phase, which inspires us to explore more FFN skipping opportunities in decoding phases. In a nutshell, \sysname explores more potential of FFN skipping in decoding phases through online importance learning, hoping to obtain a larger speedup without losing performance.

\textbf{Insight.}
\textit{The IO similarity information of the current context can be used to explore extra FFN skipping opportunities in decoding phases.}
We find that using the initial few tokens in the decoding phase can well hit the important layers of subsequent inference, and the hit rate gradually increases with the increase of the initial window. We test LLaMA3.1-8B-128k on TREC and GovReport datasets~\cite{bai2023longbench}. For each context, we use the initial $n$ tokens in decoding phases to calculate the average IO similarity, and then observe the hit rate for subsequent decoding of the current sequence. The results are shown in Table~\ref{tab:window-size}. As the window size $n$ increases, the hit rate increases and gradually becomes constant, suggesting it is unnecessary to increase $n$ infinitely.

\begin{table*}[t]
\centering
\setlength{\tabcolsep}{1mm}
\small  % 9pt
\begin{tabular}{@{}cccc|cccc|ccc@{}}
\toprule
\multirow{4}{*}{\begin{tabular}[c]{@{}c@{}}\# Target\\Skip\\ Sublayer\end{tabular}} & \multirow{4}{*}{\begin{tabular}[c]{@{}c@{}}Theo.\\Speedup\\(SU)\end{tabular}} & \multirow{4}{*}{Model} & \multirow{4}{*}{\begin{tabular}[c]{@{}c@{}}Skipping\\ Strategy\end{tabular}} & \multicolumn{4}{c|}{Prefilling Layer-Skipping} & \multicolumn{3}{c}{Decoding Layer-Skipping} \\ \cmidrule{5-11} 
 &  &  &  & \multicolumn{1}{c|}{Doc QA} & \multicolumn{2}{c|}{Few-shot Learning} & \multirow{3}{*}{\begin{tabular}[c]{@{}c@{}}Actual\\ SU\end{tabular}} & \multicolumn{2}{c|}{Text Summarization} & \multirow{3}{*}{\begin{tabular}[c]{@{}c@{}}Actual\\SU\end{tabular}} \\ \cmidrule{5-7} \cmidrule{9-10}
 &  &  &  & \multicolumn{1}{c|}{MFieldQA} & TriviaQA & \multicolumn{1}{c|}{TREC} &  & \multicolumn{1}{c|}{GovReport} & \multicolumn{1}{c|}{MultiNews} &  \\
 &  &  &  & \multicolumn{1}{c|}{(F1)} & (F1) & \multicolumn{1}{c|}{(ACC)} &  & \multicolumn{1}{c|}{(Rouge-L)} & \multicolumn{1}{c|}{(Rouge-L)} &  \\ 
\midrule
\multirow{3}{*}{0} & \multirow{3}{*}{1.00} & LLaMA3.1-8B-128k & Full Model & 29.7 & 91.6 & 75.0 & 1.00 & 34.2 & 25.8 & 1.00\\
 &  & InternLM-7B-8k & Full Model & 26.6 & 70.4 & 50.4 & 1.00 & 18.2 & 17.6 & 1.00\\
 &  & Vicuna-v1.5-7B-16k & Full Model & 32.9 & 87.8 & 68.9 & 1.00 & 27.2 & 22.4 & 1.00\\ \midrule
\multirow{12}{*}{8} & \multirow{12}{*}{1.14} & \multirow{4}{*}{LLaMA3.1-8B-128k} & Early Exit & 13.1 & 18.5 & 28.3 & \textbf{1.10} & 16.8 & 14.5 & 1.11\\
 &  &  & SkipDecode & 0.4 & 0.0 & 0.0 & \textbf{1.10} & 19.3 & 16.3 & 1.07\\
 &  &  & Unifed   Skipping & 3.4 & 4.7 & 2.2 & \textbf{1.10} & 28.2 & 22.8 & 1.11\\
 &  &  & AdaSkip & \textbf{23.4} & \textbf{86.6} & \textbf{72.8} & 1.09 & \textbf{30.9} & \textbf{24.0} & \textbf{1.15}\\ \cmidrule{3-11} 
 &  & \multirow{4}{*}{InternLM-7B-8k} & Early Exit & 6.1 & 28.2 & 32.8 & 1.13 & 3.1 & 3.6 & 1.12\\
 &  &  & SkipDecode & 0.0 & 0.0 & 0.0 & 1.13 & 11.0 & 10.6 & 1.08\\
 &  &  & Unifed   Skipping & 15.4 & 21.1 & 13.3 & 1.13 & 9.8 & 10.1 & 1.12\\
 &  &  & AdaSkip & \textbf{23.9} & \textbf{60.3} & \textbf{42.7} & \textbf{1.25} & \textbf{13.7} & \textbf{13.3} & \textbf{1.24}\\ \cmidrule{3-11} 
 &  & \multirow{4}{*}{Vicuna-v1.5-7B-16k} & Early Exit & 18.5 & 73.7 & 29.4 & 1.11 & 12.2 & 13.3 & 1.13\\
 &  &  & SkipDecode & 0.0 & 0.0 & 0.0 & 1.11 & 4.1 & 4.5 & 1.07\\
 &  &  & Unifed   Skipping & 0.0 & 0.0 & 0.0 & 1.09 & 2.6 & 2.3 & 1.12\\
 &  &  & AdaSkip & \textbf{29.6} & \textbf{82.4} & \textbf{66.1} & \textbf{1.15} & \textbf{23.6} & \textbf{20.2} & \textbf{1.20}\\ \midrule
\multirow{12}{*}{16} & \multirow{12}{*}{1.33} & \multirow{4}{*}{LLaMA3.1-8B-128k} & Early Exit & 11.4 & 4.5 & 7.8 & \textbf{1.23} & 4.9 & 5.3 & 1.26\\
 &  &  & SkipDecode & 0.0 & 0.1 & 0.0 & \textbf{1.23} & 15.2 & 13.8 & 1.15\\
 &  &  & Unifed   Skipping & 0.6 & 1.0 & 0.0 & 1.22 & 12.0 & 8.7 & 1.26\\
 &  &  & AdaSkip & \textbf{18.0} & \textbf{62.3} & \textbf{72.2} & 1.22 & \textbf{17.5} & \textbf{19.1} & \textbf{1.32}\\ \cmidrule{3-11} 
 &  & \multirow{4}{*}{InternLM-7B-8k} & Early Exit & 0.7 & 0.5 & 6.1 & 1.31 & 0.6 & 0.4 & 1.28\\
 &  &  & SkipDecode & 0.0 & 0.0 & 0.0 & 1.31 & 9.2 & 9.8 & 1.16\\
 &  &  & Unifed   Skipping & 5.1 & 0.4 & 5.0 & 1.31 & 5.7 & 6.4 & 1.28\\
 &  &  & AdaSkip & \textbf{17.2} & \textbf{38.7} & \textbf{29.4} & \textbf{1.51} & \textbf{9.4} & \textbf{9.8} & \textbf{1.47}\\ \cmidrule{3-11} 
 &  & \multirow{4}{*}{Vicuna-v1.5-7B-16k} & Early Exit & 9.6 & \textbf{41.4} & 15.0 & 1.25 & 3.0 & 3.6 & 1.28\\
 &  &  & SkipDecode & 0.0 & 0.0 & 0.0 & 1.25 & 4.8 & 3.8 & 1.16\\
 &  &  & Unifed   Skipping & 0.0 & 0.0 & 0.0 & 1.25 & 2.3 & 2.2 & 1.28\\
 &  &  & AdaSkip & \textbf{10.6} & 39.0 & \textbf{43.9} & \textbf{1.31} & \textbf{13.7} & \textbf{14.7} & \textbf{1.40}\\ \bottomrule
\end{tabular}
\caption{Evaluation of different skipping strategies.}
\label{tab:BigRes}
\end{table*}
\textbf{Method.} \label{porcedure} 
Based on the above insight, the major workflow of online importance learning mainly consists of the similarity learned from the decoding phase of the new inference task. 
Specifically, for the new context, we define the first $P$ decoded tokens as \textit{online learning windows}. These tokens are processed with unskipped layers in order to obtain current decoding features. We denote the set of FFN sublayers to be skipped in decoding phases by $skipped^P$. For $j$-th sublayer, given the input and output vectors of $t$-th decoded token as $\vec{a}^j_{t}$ and $\vec{b}^j_{t}$, $\bar{\textit{Similarity}_j}$ of $j_{th}$ FFN sublayer for current context from the first decoded token to $P_{th}$ tokens for FFN sublayers can be formulated as: 
\begin{equation}
    \bar{\textit{Similarity}}^P_j = \frac{\sum^{P}_{t=1}\textit{Similarity}(\vec{a}^j_{t}, \vec{b}^j_{t})}{P}
\end{equation} 

We get all indexes of FFN sublayers, i.e. $\textit{index}$, and the indexes of all the layers skipped in the prefilling phase, i.e., $\textit{skipped}$. To find out which layers in $\textit{index}$ set need to be skipped, we derive a threshold $\beta$ by observing the $\textit{skipped}$ set and then use this threshold to filter the additional skipped FFN sublayers. The threshold $\beta$ is the least $\textit{Similarity}$ value in $\textit{skipped}$, i.e. $\beta = \min \{ \bar{\textit{Similarity}_j} \mid j \in \textit{skipped} \}$.

We then traverse $\textit{index}$ to find the sublayers whose $\bar{\textit{Similarity}^P_j}$ is above $\beta$, and these sublayers are the additional ones to be skipped in the new context. By combining the indexes of these sublayers with the indexes of the $\textit{skipped}$ set, we obtain the adaptive sublayer-wise skipping set, denoted as $\textit{skipped}^P$.
At last, similar to the last section, we use $\bar{\textit{Scale}}_j$ to compensate for the potential deviation.

\section{Experiments}
\label{sec:exp}
In this section, we thoroughly evaluate the performance of \sysname in long-context inference. We first show the experiment settings including the benchmarks, baselines, and setups. Then we show the experiment results and analysis. 

\subsection{Experiment Settings}\label{sec:exp:setup}

\subsubsection{Benchmarks}
We select benchmarks based on the most representative long-context application scenarios \cite{bai2024longbenchbilingualmultitaskbenchmark}, encompassing tasks such as document QA, few-shot learning, and summarization.
In order to better evaluate the performance of different skipping strategies in prefilling and decoding phases, we divide the benchmarks into two sets, \textit{prefilling tasks} and \textit{decoding tasks}, according to the average output length. Specifically, we select MultiFieldQA~\cite{bai2023longbench}, TriviaQA~\cite{joshi2017triviaqa}, and TREC~\cite{li-roth-2002-learning} as prefilling tasks, with average input lengths of 6493, 8677, and 8208, respectively, and output lengths capped at 32. For decoding tasks, we choose GovReport~\cite{huang2021efficient} and MultiNews~\cite{fabbri2019multi}, with average input lengths of 9214 and 8265, and output lengths limited to 512. At last, we evaluate the end-to-end performance of all layer-wise skipping strategies by skipping layers in both prefilling and decoding phases.

\subsubsection{Baselines and Setups.} \label{sec:exp:setup:baseandsetups}
Three layer-wise skipping strategies are considered as baselines: (1) SkipDecode~\cite{del2023skipdecode} skips the initial layers except for the first one, representing early skipping; (2) Unified Skipping~\cite{liu2024accelerating} uniformly skips the intermediate layers except for the first and last layers, representing periodic skipping; and (3) Early Exit~\cite{varshney2023accelerating, fan2024not} skips the last few layers. Note that layer-wise skipping skips two sublayers, i.e., attention and FFN, in a single layer-skip operation. Specifically, as these baselines were originally designed to skip layers only during the decoding phase, we limit layer skipping to the decoding phase in decoding tasks to ensure a fair comparison. Three of the latest and widely adopted long-context LLMs are tested: LLaMA3.1-8B-128k, InternLM-7B-8k, and Vicuna-v1.5-7B-16k. A single L20 GPU with CUDA version 12.1 is used as the testbed.

\subsection{Results of Prefilling Tasks}
The middle of Table~\ref{tab:BigRes} presents the results of the prefilling tasks. Given the same number of target skip sublayers, \sysname significantly outperforms the other baselines in both Doc QA and Few-shot Learning tasks. For example, on the LLaMA3.1-8B-128k model, with a target skip sublayer number of 8, \sysname achieves a classification accuracy of 72.8\% on TREC and an F1 score of 86.6 on TriviaQA, closely approximating the performance of the full model. Even with up to 16 skipped sublayers, \sysname's accuracy on TREC remains at 72.2\%. In contrast, the accuracy of the SkipDecode and Unified Skipping approaches decrease by more than 90\% when skipping only 8 sublayers (4 whole layers).

The significant disparity between the baselines and \sysname underscores the critical importance of identifying the distribution of layer significance for accuracy in long-context prefilling tasks. Fixed layer-skipping strategies fail to adapt effectively across different models.

In terms of speedup, the computational complexity of attention scales quadratically with sequence length, making attention computations more demanding than those of FFN in long-context scenarios. Due to \sysname skipping more attention sublayers, it achieves over a 10\% speedup advantage on InternLM compared to the baseline. For the LLaMA model, the attention sequence parallelism and other optimization techniques are relatively mature, making the FFN execution time longer during the prefilling phase. As a result, our approach is slightly outperformed by the baseline.

\begin{table*}[t]
\centering
\setlength{\tabcolsep}{1mm}
\small  % 9pt
\begin{tabular}{cc|cc|cc|cc}
\toprule
\multirow{3}{*}{\begin{tabular}[c]{@{}c@{}}\#   Target\\Skip\\      Sublayer\end{tabular}} & \multirow{3}{*}{\begin{tabular}[c]{@{}c@{}}Skipping\\      Strategy\end{tabular}} & \multicolumn{2}{c|}{LLaMA-3.1B-128k} & \multicolumn{2}{c|}{InternLM-7B-8k} & \multicolumn{2}{c}{Vicuna-v1.5-7B-16k} \\ \cmidrule{3-8} 
 &  & GovReport & MultiNews & GovReport & MultiNews & GovReport & MultiNews \\
 &  & (Rouge-L) & (Rouge-L) & (Rouge-L) & (Rouge-L) & (Rouge-L) & (Rouge-L) \\ \midrule
0 & Full Model & 34.2 & 25.8 & 18.2 & 17.6 & 27.2 & 22.4 \\ \midrule
\multirow{4}{*}{8} & Early Exit & 15.3 & 12.4 & 2.7 & 3.4 & 12.2 & 13.3 \\
 & SkipDecode & 1.0 & 1.1 & 0.0 & 0.0 & 0.0 & 0.0 \\
 & Unifed Skipping & 1.6 & 1.1 & 8.9 & 9.9 & 0.0 & 0.0 \\
 & AdaSkip & \textbf{30.5} & \textbf{24.1} & \textbf{12.9} & \textbf{12.7} & \textbf{23.0} & \textbf{21.1} \\ \midrule
\multirow{4}{*}{16} & Early Exit & 4.3 & 4.4 & 0.5 & 0.3 & 1.9 & 1.8 \\
 & SkipDecode & 0.0 & 0.0 & 0.0 & 0.0 & 0.0 & 0.0 \\
 & Unifed  Skipping & 0.0 & 0.1 & 0.4 & 1.0 & 0.0 & 0.0 \\
 & AdaSkip & \textbf{18.9} & \textbf{17.8} & \textbf{7.7} & \textbf{7.1} & \textbf{13.6} & \textbf{14.1} \\ \bottomrule
\end{tabular}
\caption{Evaluation on End-to-End Skipping Strategies.}
\label{tab:PD_Skip}
\end{table*}
\subsection{Results of Decoding Tasks}
The right half of Table~\ref{tab:BigRes} presents the results of decoding tasks. Despite the baselines being specifically tailored for decoding tasks, our method consistently demonstrates superior performance. Even with the number of skipped sublayers reaching 16, we still maintain comparable performance. For instance, the LLaMA model achieves Rouge-L scores exceeding 17.5 on both datasets, comparable to the full InternLM model.
%The Vicuna model's Rouge-L scores also exceed 13, outperforming the baselines by approximately 10 points.
It is noteworthy that the Early Exit method, which performs reasonably well in the prefilling tasks, fails to maintain generation quality during decoding. Its Vicuna Rouge-L scores for the two summarization tasks fall below 4.0, possibly due to the accumulation of errors in the autoregressive process. In contrast, \sysname accurately identifies the least significant sublayers, allowing LLM to maintain optimal performance even with additional skipping of FFNs.

In terms of execution speed, the inference time during the decoding phase is primarily dictated by HBM access. In long-context inference, attention operates slower than FFN due to the extensive KV cache access required. Our approach achieves a higher acceleration ratio by skipping more attention layers at the outset, thanks to the higher attention similarity obtained during the offline learning phase. After online learning, we further enhance the acceleration by selectively skipping additional FFN layers. Overall, our method delivers up to a 17\% acceleration improvement compared to the baseline. The Skip Decode approach achieves the lowest speedup because it employs a progressive layer skipping strategy, where the number of skipped layers gradually increases with the decoding steps until reaching the preset number.

\subsection{Results of End-to-End Testing} \label{sec:exp:full}
We evaluate the end-to-end performance of various layer skipping strategies, namely, implementing simultaneous skipping during both the prefilling and decoding phases. As demonstrated by Table~\ref{tab:PD_Skip}, the performance of existing approaches significantly degrades when layer skipping is applied in both phases, compared to applying it solely during decoding. The SkipDecode approach causes Rouge-L scores to plummet to nearly zero across all three models. Similarly, Unified Skipping, which previously exhibited a modest difference from our approach on specific data points in decoding tasks, sees all its Rouge-L scores drop below 10.0 in this scenario. Additionally, when skipping 16 sublayers, the Early Exit approach yields scores below 5.0 across all models.

The results highlight the significant limitations of existing methods, which are unable to effectively apply layer skipping during both the prefilling and decoding phases in tasks with longer generation lengths. In contrast, our approach maintains nearly identical performance as when layer skipping is applied only during the decoding phase, demonstrating that our skipping strategy effectively adapts to both the prefilling and decoding phases. In real-world long-context tasks, our method exhibits exceptional practical value due to its ability to employ a complete layer-skipping strategy. It can markedly optimize the TTFT introduced during the prefilling phase and reduce the storage costs of the KV cache for long prompts.

\section{Related Work}

\textbf{Long-context Model.}
With the growing demand for long-context models, numerous studies have concentrated on expanding the context window of LLMs. Many models have fine-tuned LLaMA-2 by scaling Rotary Position Embeddings (RoPE)~\cite{su2023enhanced}, expanding its input window to 32k, as seen in LongChat~\cite{li2023long}, and to 128k, as demonstrated in Yarn-LLaMA-2~\cite{peng2023yarn}. By leveraging length extrapolation, the context windows can extend beyond 1 million tokens~\cite{liu2023scaling}. However, these approaches do not alleviate the substantial inference costs associated with long-context processing.

\noindent\textbf{Long-context LLM Inference Optimization.} Given the substantial increase in KV cache size introduced by long sequences, many studies have concentrated their inference optimization efforts on compressing, evicting, and reusing KV cache. Heavy Hitter Oracle (H2O)~\cite{zhang2024h2o} retains a limited budget of the important KV cache based on the sum of historical attention scores. SnapKV~\cite{li2024snapkv} reduces memory access during decoding by observing the attention distribution of the prompt's tail over the prefix to selectively filter the corresponding KV cache, thereby achieving acceleration. PyramidKV~\cite{zhang2024pyramidkv} optimizes KV cache storage more flexibly by allocating different KV cache budgets to various layers and attention heads based on the observed information flow aggregation patterns. However, these approaches fail to address the substantial computational burden associated with generating extensive KV cache during the long sequence prefilling stage.
\section{Conclusion} 
In conclusion, this paper focuses on exploring the layer-wise skipping strategy in long-context inference. It first discusses the typical challenges in long-context inference and presents a detailed examination of the importance distribution of various components including layer and sublayer modules such as attention and FFN across a variety of different models. The analysis underlines the limitations of the current layer-wise skipping strategies in long-context inference. In response to these limitations, this paper proposes a novel, auto-adaptive, sublayer-wise skipping strategy that requires no training and is applicable to both the prefilling and decoding phases. Through rigorous testing across a diverse array of long-context datasets and models, we have demonstrated that our system, \sysname, significantly outperforms the baseline in both generation quality and inference speed.

\section{Acknowledgments}
This work was supported in part by China NSF grant No. 62202297, Open Project Program of Laboratory of Pinghu, and Huawei Cloud. The opinions, findings, conclusions, and recommendations expressed in this paper are those of the authors and do not necessarily reflect the views of the funding agencies or the government.

\end{document}